\documentclass{acm_proc_article-sp}
\usepackage{graphicx}
\usepackage{graphics}
\usepackage{epstopdf}
\usepackage{amsmath,amsfonts,amssymb}
\usepackage{algorithm}
\usepackage{algpseudocode}
\usepackage{hyperref}
\usepackage{color}
\begin{document}

\title{Multi-Task Metric Learning on Network Data}
%\subtitle{[Extended Abstract]
%\titlenote{A full version of this paper is available as
%\textit{Author's Guide to Preparing ACM SIG Proceedings Using
%\LaTeX$2_\epsilon$\ and BibTeX} at
%\texttt{www.acm.org/eaddress.htm}}}
%
% You need the command \numberofauthors to handle the 'placement
% and alignment' of the authors beneath the title.
%
% For aesthetic reasons, we recommend 'three authors at a time'
% i.e. three 'name/affiliation blocks' be placed beneath the title.
%
% NOTE: You are NOT restricted in how many 'rows' of
% "name/affiliations" may appear. We just ask that you restrict
% the number of 'columns' to three.
%
% Because of the available 'opening page real-estate'
% we ask you to refrain from putting more than six authors
% (two rows with three columns) beneath the article title.
% More than six makes the first-page appear very cluttered indeed.
%
% Use the \alignauthor commands to handle the names
% and affiliations for an 'aesthetic maximum' of six authors.
% Add names, affiliations, addresses for
% the seventh etc. author(s) as the argument for the
% \additionalauthors command.
% These 'additional authors' will be output/set for you
% without further effort on your part as the last section in
% the body of your article BEFORE References or any Appendices.

\numberofauthors{2} %  in this sample file, there are a *total*
% of EIGHT authors. SIX appear on the 'first-page' (for formatting
% reasons) and the remaining two appear in the \additionalauthors section.
%
\author{
% You can go ahead and credit any number of authors here,
% e.g. one 'row of three' or two rows (consisting of one row of three
% and a second row of one, two or three).
%
% The command \alignauthor (no curly braces needed) should
% precede each author name, affiliation/snail-mail address and
% e-mail address. Additionally, tag each line of
% affiliation/address with \affaddr, and tag the
% e-mail address with \email.
%
% 1st. author
\alignauthor
\mbox{Chen Fang}\\
       \affaddr{Department of Computer Science }\\
       \affaddr{Dartmouth College}\\
       \affaddr{Hanover, NH, 03755, U.S.A.}\\
       \email{chenfang@cs.dartmouth.edu}
%% 2nd. author
%\alignauthor
%       \mbox{\affaddr{}}
%       \affaddr{}
%       \affaddr{}
%       \affaddr{}
%       \email{}
\alignauthor
Daniel N. Rockmore\\
       \affaddr{Department of Computer Science}\\
       \affaddr{Department of Mathematics}\\
       \affaddr{Dartmouth College}\\
       \affaddr{Hanover, NH, 03755}\\
       \email{rockmore@cs.dartmouth.edu}
}
% There's nothing stopping you putting the seventh, eighth, etc.
% author on the opening page (as the 'third row') but we ask,
% for aesthetic reasons that you place these 'additional authors'
% in the \additional authors block, viz.
%\additionalauthors{Additional authors: John Smith (The Th{\o}rv{\"a}ld Group,
%email: {\texttt{jsmith@affiliation.org}}) and Julius P.~Kumquat
%(The Kumquat Consortium, email: {\texttt{jpkumquat@consortium.net}}).}
%\date{30 July 1999}
% Just remember to make sure that the TOTAL number of authors
% is the number that will appear on the first page PLUS the
% number that will appear in the \additionalauthors section.
\maketitle
\begin{abstract}
Multi-task learning (MTL) has been shown to improve prediction performance in a number of different contexts by learning models jointly on multiple different, but related tasks. Network data, which are a priori data with a rich relational structure, provide an important context for applying MTL. 
%Due to the diversity and variation in network data, various tasks can be performed and lots of them are \emph{correlated}. Thus, it is important and valuable to do MTL on networks. 
In particular, the explicit relational structure implies that network data is not i.i.d. data. Network data also often comes with significant metadata (i.e., attributes) associated with each entity (node). Moreover, due to the diversity and variation in network data (e.g., multi-relational links or multi-category entities), various tasks can be performed and often a rich correlation exists between them. Learning algorithms should exploit all of these additional sources of information for better performance. In this work we take a metric-learning point of view for the MTL problem in the network context. Our approach builds on  {\em structure preserving metric learning} (SPML)~\cite{ShawHuangJebara2011}.  In particular SPML learns a Mahalanobis distance metric for node attributes using network structure as supervision, so that the learned distance function encodes the structure and can be used to predict link patterns from attributes. In the fundamental paper  \cite{ShawHuangJebara2011} SPML is described for single-task learning on single network. Herein, we propose a {\em multi-task} version of SPML, abbreviated as MT-SPML, which is able to learn across multiple related tasks on multiple networks via shared intermediate parametrization. MT-SPML learns a specific metric for each task and a common metric for all tasks. The task correlation is carried through the common metric and the individual metrics  encode task specific information. When combined together, they are structure-preserving with respect to individual tasks. MT-SPML works on general networks, thus is suitable for a wide variety of problems. In experiments, we challenge MT-SPML with two common real-word applications: citation prediction for Wikipedia articles and social circle prediction in Google+. Our results show that MT-SPML achieves significant improvement over other competing methods.\\
\end{abstract}
%\vspace{2mm}
% A category with the (minimum) three required fields
%A category including the fourth, optional field follows...
%\category{D.2.8}{Software Engineering}{Metrics}[complexity measures, performance measures]
%\vspace{-4mm}
%\vspace{-4mm}
 % NOT required for Proceedings
\section{Introduction}
Multi-task learning (MTL) \cite{Evgeniou_2004, Caruana97multitasklearning, Yu_2005, Agarwal_2010, Daume2009BML} considers the problem of learning models jointly and simultaneously over multiple, different but related tasks. Compared to single-task learning (STL), which learns a model for each task independently using only task specific data, MTL leverages all available data and shares knowledge among tasks, thereby resulting in better model generalization and prediction performance. The underlying principle of MTL is that highly correlated tasks can benefit from each other via joint training, but additional care should be taken to respect the distinctiveness of each task, i.e., it is usually inappropriate to pool all available data and learn a single model for all tasks.  

Despite the popularity and value of MTL, most MTL methods are developed for tasks on i.i.d. data. Standard examples include phoneme recognition~\cite{Seltzer_deep2013} and image recognition~\cite{Wang_cvpr09}. Explicitly correlated data, often represented in the form of a network, provide a rich source of new applications contexts wherein the explicit relatedness of the data might be leveraged to improve performance on similarly correlated tasks.  \textit{That is, although each task bears its own distinctiveness, relatedness  cannot be ignored and should be exploited for good!} The following two scenarios, provide two important examples where it is beneficial to exploit the correlation between tasks. These scenarios are in fact the settings for the experiments using real-world data that we present in   Section 4.

\textbf{Scenario 1: Article citation prediction}\newline
Articles tend to cite each other, especially those in the same subject area. The citation prediction problem has been studied extensively~\cite{Hasan06linkprediction, Liben_Nowell_2003, Taskar03linkprediction, Adafre:2005:DML:1134271.1134284, 4053061}. People either build a predictive model for a unified network~\cite{Liben_Nowell_2003} (i.e., a citation network that contains papers of all subject areas,) or build predictive models for each area independently~\cite{ShawHuangJebara2011}. Since article content and citation pattern varies across different areas, the former methodology ignores the difference between areas. However, some areas, while labeled as different are still related, in the sense of both their  content and citation pattern. Thus the latter methodology fails to exploit the correlation among subject areas. For example, computer science and electrical engineering articles may be classified or tagged as different areas, but in many cases they may still have much in common, or at least have significant similarity or overlap. In this case, to build predictive models for citations, a learning algorithm that is capable of utilizing these overlaps and explicit commonalities 
%knowledge among different areas 
has advantages over traditional methods.

\textbf{Scenario 2: Social circle prediction}\newline
Members of online social networks tend to categorize their links to followers/followees. For example, many social networking platforms enable coarse-scale categorizations such as  ``family members," or ``friends and colleagues." Finer gradations allow for categorizations such as colleagues at particular companies or classmates at specific schools. A person's {\em social circle}, studied in~\cite{nips_McAuleyL12}, is the ego network of a social network user (or ``ego") in which all links belong to the same category.  I.e.,  the induced subgraph of a given entity containing only links of a given type. Given a friend or stranger, the goal of social circle prediction is to assign him/her to appropriate social circles. Because some social circles are related to each other (e.g., family members and childhood friends may share some common informative features such as geological proximity), advantages may very well accrue if the relatedness of the entities was used for the various predictions,  instead of building predictive assignment model for each social circle independently.

\medskip

As these scenarios suggest,  there should be advantages to developing methods that can leverage the correlations among tasks on network data. In what follows we show that MTL is a natural direction to pursue and that it does in fact provide some significant improvements. 

Different from i.i.d. data, network data not only has attributes (metadata) associated with each entity (node), but rich structural information, mainly encoded in the links. Both attributes and structure should be exploited in learning. Structure preserving metric learning (SPML), originally developed for single-task learning~\cite{ShawHuangJebara2011} is such a method. It learns a Mahalanobis distance metric for node attributes by using network structure as supervision, so that the learned distance function encodes the structure and can be used to predict link patterns from attributes. Inspired by the use of SPML in the single task context, we propose its multi-task version, MT-SPML, which learns Mahalanobis distance metrics jointly over all tasks. More precisely, it learns a common metric for all tasks and one metric for each individual task. The common metric construction follows the methodology of shared intermediate parameterization~\cite{Evgeniou_2004,parameswaran10large}, which allows sharing knowledge between tasks. While a single task specific metric captures task specific information, when combined, they work together to preserve the connectivity structure of the corresponding network, thus are useful for link prediction from attributes. We further show that as in the case of SPML, MT-SPML can be optimized with efficient online methods similar to OASIS~\cite{ChechikSSB10} and PEGASOS~\cite{Shalev_ShwartzSSC11} via stochastic gradient descent. Finally, MT-SPML is designed for general networks, and in experiments we apply MT-SPML to two common, but different real-world prediction problems (citation prediction and social circle prediction) with promising results.

\section{Related Work}
There is a large body of work on MTL for i.i.d. data. Yu et al.~\cite{Yu_2005} applied hierarchical Bayesian modeling to nonparametric Gaussian processes, and the resulting method was used for text categorization. Evgeniou et al.~\cite{Evgeniou_2004} extended Support Vector Machines (SVMs) to MTL via parameter sharing, and the method was applied to learn predictive models for exam scores of student at different schools. Following the same intuition as~\cite{Evgeniou_2004}, Parameswaran et al.~\cite{parameswaran10large} proposed the multi-task version of large margin nearest neighbor metric learning~\cite{WeinbergerS09}, which was tested on speech recognition. In~\cite{Wang_cvpr09}, Wang et al. applied MTL to help face recognition and image retrieval. Very recently, Seltzer et al.~\cite{Seltzer_deep2013} showed  how multi-task deep neural network can further help phoneme recognition. 

Researchers also have been studying the problem of learning across multiple graph data for various purposes. Zhou et al.~\cite{ZhouZYSTZG08} improved document recommendation by finding an embedding for multiple graphs via matrix factorization. In~\cite{icdm_TangLD09}, Tang et al. attempted to do clustering jointly over different graphs. Prakash et al.~\cite{cikm_ComarTJ10} developed an algorithm to jointly do clustering and classification on networks. In the area of relational learning, tensor decomposition-based methods are usually applied~\cite{journals/corr/abs-1303-1733} for problems on multi-relational data.

Of greatest relevance for our work is~\cite{icde_QiAH13}, wherein Qi et al. carefully developed a mechanism to sample across networks to predict missing links in a target network. Our paper differs from it in (at least) the following ways. First and foremost, we aim at improving prediction performance of all networks, while~\cite{icde_QiAH13} targets at a specific network and uses sources networks to help link prediction on it. Second, MT-SPML essentially learns a joint embedding of both attribute features and network topological structure, while~\cite{icde_QiAH13} tries to linearly combine attribute features and local structure information, e.g. the number of shared neighbors between a pair of nodes. This is another important difference. In this paper, we are looking at improving the performance of predicting links only from attributes. Hence, given an incoming node at test time, we do not have any prior knowledge of its connectivity information, which is not an uncommon scenario in practice, e.g. nascent network or sparse network. In these scenarios,~\cite{icde_QiAH13} is more likely to waste its modeling power on local structure features, which is often unavailable. The lack of initial structure information also makes our problem somewhat more difficult than the traditional link prediction problem, which has a snapshot of current network, which is usually not sparse, and predicts future links among already observed nodes. Nevertheless, the metric learned by our approach can help the traditional link prediction as well if attributes are available. Last but not least, we aim to naturally marry MTL and network/graph/relational data, to take advantage of MTL's ability of handling relatedness and heterogeneity. The proposed MT-SPML is a general method, which can handle different types of correlations and variations among tasks (e.g.,the marginal distribution of node attributes differs from task to task, and the semantics or types of links can also vary depending on specific task). Thus, our approach can be applied to general network data, like article citation, social networks and email networks.

\section{Our Approach}
In this section, we will first cover the technical details of Structure Preserving Metric Learning (SPML). Then, both derivation and sketched proof of MT-SPML are provided.

\subsection{Notations and preliminaries}
Given a network, we model it as a graph $\textbf{G}=(\textbf{X},\textbf{A})$, where $\textbf{X} \in \mathbb{R}^{d \times n}$ represents the node attributes and $\textbf{A} \in \mathbb{R}^{n \times n}$ is the binary adjacency matrix, whose entry $\textbf{A}_{ij}$ indicates the linkage information between node $i$ and node $j$. A Mahalanobis distance is parameterized by a positive semidefinite (PSD) matrix $\textbf{M} \in \mathbb{R}^{d \times d}$, where $\textbf{M} \succeq 0$. The corresponding distance function is define as $d_\textbf{M}(x_i,x_j)= {(x_i-x_j) ^\top \textbf{M}(x_i-x_j)}$. This is equivalent to the existence of a  linear transformation $\textbf{L} \in \mathbb{R}^{d \times d}$ on the feature space such that $\textbf{M}=\textbf{L}^\top\textbf{L}$. Given a metric $\mathbf{M}$, to predict the structure pattern of $\textbf{X}$, we adopt the simple $k$-nearest neighbor algorithm, which is denoted as $\mathcal{C}$, meaning each node is connected with its top-$k$ nearest neighbors under the defined metric. Mathematically, we denote it as $\mathcal{C}(\textbf{X},\textbf{M})=\textbf{A}$, and we say $\textbf{M}$ is {\em structure preserving} or that it {\em preserves \textbf{A}}. The Frobenius norm is demoted as $||\cdot||_F$.

We denote a set of networks as $\mathcal{G}=\{\textbf{G}_1,\textbf{G}_2,\ldots,\textbf{G}_Q\}$. Each individual network $\textbf{G}_q$ has its own $\textbf{X}_q$ and $\textbf{A}_q$. We use $q$ to index each network and $(i,j)$ for its element. Thus,  for notational simplicity $\textbf{A}_{qij}$ stands for element $(i,j)$ in $\textbf{A}_q$. Similarly, $X_{qi}$ represents the feature of node $i$ in $\textbf{X}_q$. In algorithms, we will use a superscript to index over iteration, e.g., $\textbf{M}^k$ refers to the $k$th iteration of $\textbf{M}$ under the relevant iterative process.

\subsection{SPML}
The goal of SPML is to learn $\textbf{M}$ from a network $\textbf{G}=(\textbf{X},\textbf{A})$, such that $\textbf{M}$ preserves $\textbf{A}$. This problem has a semidefinite max margin learning formulation, which is as follows:
\begin{equation}
\min_{\textbf{M} \succeq 0} \frac{\lambda}{2}||\textbf{M}||{^2_F}+\xi
\end{equation}
subject to the following constraints:
\begin{equation}
\forall_{i,j}, \hspace{2mm} d_{\textbf{M}}(x_i,x_j)\geq(1-\textbf{A}_{ij})\max_l(\textbf{A}_{il}d_{\textbf{M}}(x_i,x_l))+1-\xi
\end{equation}
In (1) the Frobenius norm is a regularizer on $\textbf{M}$ and $\lambda$ is the corresponding weight parameter. The key idea to achieve structure preserving is the set of linear constraints in (2). This  essentially enforces that from node $i$, the distances to all disconnected nodes must be larger than the distance to the furthest connected node. Thus, when the constraints in (2) are all satisfied, $\mathcal{C}(\textbf{X},\textbf{M})$ will exactly produce $\textbf{A}$. Furthermore, to allow for violation (with penalty), the slack variable $\xi$ is introduced to both (1) and (2).

With the many  constraints in (2), optimizing (1) becomes unfeasible when the network has few hundreds of nodes. But a  rewriting of the problem as follows allows us to use stochastic subgradient descent  (see Algorithm 1):
\begin{equation}
f(\textbf{M})=\frac{\lambda}{2}||\textbf{M}||^2_F+\frac{1}{|S|}\sum_{(i,j,l)\in S}\max(\Delta_\textbf{M}(x_i,x_j,x_l)+1,0)
\end{equation}
where $\Delta_\textbf{M}(x_i,x_j,x_l) = d_\textbf{M}(x_i,x_l)-d_\textbf{M}(x_i,x_j)$. $S$ is defined as $S=\{(i,j,l)|\textbf{A}_{i,l}=1  \land \textbf{A}_{i,j}=0\}$, so the triplet $(i,j,l)$ means that there is a link between node $i$ and node $l$, but not between $i$ and $j$. The subgradient of  (3) can be calculated as
\begin{multline}
\bigtriangledown f=\lambda \textbf{M}+\frac{1}{|S|}\sum_{(i,j,l)\in S_+}\hspace{-3mm}\biggl(\left(x_i-x_l \right)(x_i-x_l)^\top \\ -(x_i-x_j)(x_i-x_j)^\top \biggr)
\end{multline}
where $S_+$ is the set of triplets whose hinge losses are positive. At every iteration $t$ of Algorithm 1, $B$ triplets are randomly sampled and the corresponding stochastic subgradient is calculated with regard to the current metric $\textbf{M}^t$ and these triplets. Since Algorithm 1 is a variant of PEGASOS~\cite{Shalev_ShwartzSSC11}, its complexity does not depend on training set size $n$, but on feature dimensionality $d$. For the number of iterations $T$ needed to reach convergence, proved by~\cite{ShawHuangJebara2011,Shalev_ShwartzSSC11}, it depends on parameter $\lambda$ and optimization error, which measures how close the final objective value is to the global optimal objective value. Notice that after updating $\textbf{M}$, there is an optional step, in which the current $\textbf{M}$ is projected to its PSD cone. Experiments in~\cite{ShawHuangJebara2011} show that delaying this operation to the end of the algorithm works well in practice and further reduces computational complexity.

\begin{algorithm}
\caption{Stochastic subgradient descent for SPML}
\textbf{Input:} $\textbf{G}=(\textbf{X},\textbf{A}),\lambda,T,B$\\
\textbf{Output:} $\textbf{M} \succeq 0$
\begin{algorithmic}[1]
\State $\textbf{M}^{0} \gets \textbf{I}^{d\times d}$  
\For {$t=1,2,\ldots,T$}
\State $\eta^t \gets \frac{1}{\lambda \times t}$
\State $s \gets \emptyset$
	\For {$b=1,2,\ldots,B$}
		\State Random sample $(i,j,l)$ from $S$
		\State $s \gets s \cup (i,j,l)$
	\EndFor
	\State $\textbf{M}^t \gets \textbf{M}^{t-1}-\eta^t\bigtriangledown f(\textbf{M}^{t-1},s)$
	\State $\textbf{M}^t \gets [\textbf{M}^t]_+$ (Optional: Project to PSD cone)
\EndFor
\State return $\textbf{M}^T$
\end{algorithmic}
\end{algorithm}
\subsection{MT-SPML}
In this section, we show how MT-SPML extends SPML to multi-task setting.

The input is a set of networks $\mathcal{G}=\{\textbf{G}_1,\textbf{G}_2,\ldots,\textbf{G}_Q\}$. Each network is $\textbf{G}_q=(\textbf{X}_q,\textbf{A}_q)$. Our approach is a general method, thus it works for both problems with or without nodes overlapping. Note that, the nodes of all networks have common feature spaces. MT-SPML treats each network as a task. It follows the idea of \emph{shared intermediate parametrization} as in~\cite{parameswaran10large} to enable knowledge transfer between tasks. The goal is to learn jointly over $\mathcal{G}$ a task specific metric $\textbf{M}_q$ for each task and a common metric $\textbf{M}_0$, through which knowledge transfers among tasks , so that the combined metric $(\textbf{M}_0+\textbf{M}_q)$ respects the structure of $\textbf{G}_q$, for all $\textbf{G}_q \in \mathcal{G}$. Thus, the distance between two nodes in $\textbf{G}_q$ is defined as
\begin{equation}
d_q(x_i,x_j)=(x_i-x_j)^\top(\textbf{M}_0+\textbf{M}_q)(x_i-x_j).
\end{equation}

MT-SPML is formulated as a regularized learning problem as follows:
\begin{equation}
\min_{\textbf{M}_0,\textbf{M}_1,\ldots,\textbf{M}_Q} \frac{\gamma_0}{2}||\textbf{M}_0-\textbf{I}||^2_F+\sum_{q=1}^Q\frac{\gamma_q}{2}||\textbf{M}_q||^2_F+\sum_{q=1}^Q\xi_q
\end{equation}
subject to the following constraints:
\begin{flalign}
\hspace*{0mm} & \forall q,i,j,:&
\end{flalign}
\hspace{5mm}$d_q(x_{qi},x_{qj})\geq(1-\textbf{A}_{qij})\max_l(\textbf{A}_{qil}d_q(x_{qi},x_{ql}))+1-\xi_q$. 

\medskip
In order to solve it, we rewrite it as following by incorporating the constraints:
\begin{equation}
\begin{split}
f(\textbf{M}_0,\textbf{M}_1,\ldots,\textbf{M}_Q)=\frac{\gamma_0}{2}||\textbf{M}_0-\textbf{I}||^2_F+\sum_{q=1}^Q\frac{\gamma_q}{2}||\textbf{M}_q||^2_F\\
+\sum_{q=1}^Q \frac{1}{|S_q|}\sum_{(i,j,l)\in S_q} \max (\Delta_{q}(x_{qi},x_{qj},x_{ql})+1,0)\\
\end{split}
\end{equation}

\begin{algorithm}
\caption{Stochastic subgradient descent for MT-SPML}
\textbf{Input:} $\mathcal{G}=\{\textbf{G}_1,\textbf{G}_2,\ldots,\textbf{G}_Q\}$, where $\textbf{G}_q=(\textbf{X}_q,\textbf{A}_q)$, \\ \hbox{}\hspace{10.5mm} $\gamma_0,\gamma_1,\ldots,\gamma_Q$, $T$, $B$\\
\textbf{Output:} $\textbf{M}_0, \textbf{M}_1, \ldots, \textbf{M}_Q \succeq 0$
\begin{algorithmic}[1]
%\State $\textbf{M}^{0}_0 \gets \textbf{I}^{d\times d}$  
\For {$q=0,1,\ldots,Q$}
	\State $\textbf{M}_q^0 \gets \textbf{I}^{d\times d}$
\EndFor

\For {$t=1,2,\ldots,T$}
\For {$q=1,2,\ldots,Q$}
\State $\eta^t_q \gets \frac{1}{\lambda \times t}$
\State $s_q \gets \emptyset$
	\For {$b=1,2,\ldots,B$}
		\State Random sample $(i,j,l)$ from $S_q$
		\State $s_q \gets s_q \cup (i,j,l)$
	\EndFor
\State $\textbf{M}^t_q \gets \textbf{M}^{t-1}_q-\eta^t_q\bigtriangledown_{\textbf{M}_q} f(\textbf{M}^{t-1}_q,s_q)$
	\State $\textbf{M}^t_q \gets [\textbf{M}^t_q]_+$ (Optional: Project to PSD cone)
\EndFor
\State $\textbf{M}^t_0 \gets \textbf{M}^{t-1}_0-\eta^t_0\bigtriangledown_{\textbf{M}_0} f(\textbf{M}^{t-1}_0, \{s_1,s_2,\ldots, s_Q\})$
\State $\textbf{M}^t_0 \gets [\textbf{M}^t_0]_+$ (Optional: Project to PSD cone)
\EndFor
\State return  $\textbf{M}^T_0, \textbf{M}^T_1, \ldots, \textbf{M}^T_Q \succeq 0$
\end{algorithmic}
\end{algorithm}
where $\Delta_{q}(x_{qi},x_{qj},x_{ql})=d_q(x_{qi},x_{ql})-d_q(x_{qi},x_{qj})$. Although (8) has more unknown variables than (3), with respect to each unknown, it is in the same form as (3). Therefore, (8) can be solved with the same stochastic subgradient descent method using partial subgradient. The partial subgradients of (8) with respect to $\textbf{M}_0$ and $\textbf{M}_q$ are
\begin{multline}
\bigtriangledown_{\textbf{M}_0}f = \gamma_0 \left(\textbf{M}_0-\textbf{I}\right)
\\+\sum_{q=1}^{Q}\frac{1}{|S_q|}\sum_{(i,j,l)\in S_{q+}}\biggl( \left(x_{qi}-x_{ql}\right)\left(x_{qi}-x_{ql}\right)^\top \\-\left(x_{qi}-x_{qj}\right)\left(x_{qi}-x_{qj}\right)^\top \biggr)
\end{multline}
and
\begin{multline}
\bigtriangledown_{\textbf{M}_q}f = \gamma_q \textbf{M}_q + \frac{1}{|S_q|}\sum_{(i,j,l)\in S_{q+}}\biggl( \left(x_{qi}-x_{ql}\right)\left(x_{qi}-x_{ql}\right)^\top \\-\left(x_{qi}-x_{qj}\right)\left(x_{qi}-x_{qj}\right)^\top \biggr).
\end{multline}
With all necessary information, the optimization algorithm is outlined in Algorithm 2. Algorithm 2 runs for $T$ iterations. Within each iteration, it does two things: (1) Randomly sample $B$ triplets for each task, so as to calculate the partial subgradient and update the corresponding unknown; (2) Calculate the partial subgradient of the common metric $\textbf{M}_0$ and update it using the $Q\times B$ triplets already sampled. Optionally, the metric matrices can be projected to their PSD cones. The analysis of Algorithm 1 still holds for Algorithm 2, thus it scales up with regard to feature dimensionality, optimization error and the parameters $\gamma_q$, but not the training set size.

\section{Experiments}
In this section, we present experimental results on real-world data. We apply MT-SPML to the two scenarios mentioned in the Introduction: article citation prediction and social circle prediction. We show that in both cases, MT-SPML can significantly improve prediction performance. \footnote{Code and data are available for download at \url{http://www.cs.dartmouth.edu/~chenfang/temp/MT_SPML/demo_code.tar.gz}}
\subsection{Citation prediction on Wikipedia}
The data is obtained from~\cite{ShawHuangJebara2011}. The articles from the following three areas are crawled from Wikipedia: search engine, graph theory and philosophy. The citations between articles within each area are also crawled. The goal is, given an article, to predict  the referencing of other articles within its area solely from its content. Therefore, at test time, no reference information of the test article is made available at all. The challenge of this problem is the fact that: (1) there is little node overlap between networks (i.e.,  an article belongs to only one area), thus the marginal distribution of node attributes $P(\textbf{X}_q)$ may vary drastically from area to area, which poses difficulty for knowledge transfer; (2) the conditional probability of structure on attributes $P(\textbf{A}|\textbf{X})$ may also vary, because some words are informative and indicative for some areas, but not for others. The statistics of this dataset is detailed in Table 1. Bag-of-words is used to capture article content and the dimensionality is 6695. The high dimensionality reduces the need to learn full matrices. Therefore, we choose to learn diagonal metric matrices. This further reduces computational complexity. We split the dataset 80\%/20\% as training and testing respectively, then fix the testing part and vary the size of training set by sampling from the training part. We end up sampling $20\%,40\%,60\%,80\%,$ and finally $100\%$ of the training part. Model selection is carried out on the sampled training set via 5-fold cross-validation. At test stage, for every test example our algorithm suggests other articles for citation according to their distances to test article. We build the receiver operator characteristic (ROC) curve for every test article, and use the average area under the curve (AUC) of the entire test set as performance measurement. We compare our results with two families of methods:

\textbf{SVM methods}
We apply SVM-based methods in order to show the importance of modeling network structure. Since SVM-based methods do not model network structure, we need to construct features to encode this piece of information. The training examples are constructed by taking the pairwise difference of the attributes between two nodes. The training labels are binary, with 1 representing the existence of a link between a pair of nodes and 0 the absence.  For a given edge, we measure its distance/length using the output of the classification score, which represents the confidence of having a link. Although the classification score is inversely proportional to the notion of distance, a simple conversion can make the two variables proportional to each other. Thus ROC and AUC can be calculated. The following specific methods are included:

\begin{itemize}
\item \textbf{ST-SVM}: This is the normal single-task SVM. An SVM is trained for each network independently. It does not explore the correlation between tasks. The model is trained and tested with LIBLINEAR~\cite{journals_jmlr_FanCHWL08}.
\item \textbf{U-SVM}: We train one SVM for all networks by pooling all data together. We use the capital letter ``U" to represent the naive strategy of data pooling. This is essentially ignoring the fact that training examples are from different tasks and treating it as a simple single task learning problem. The model is also trained and tested using LIBLINEAR~\cite{journals_jmlr_FanCHWL08}.
\item\textbf{MT-SVM}: This is the multi-task SVM in~\cite{Evgeniou_2004}. Similar to our model, it jointly learns a common decision boundary for all and a specific boundary for each task. At   test time, the common and task specific decision boundary together form the final classification model for each task. This method exploits task correlations via intermediate parameter sharing, but does not use network structure at the model level. We used the software from~\cite{conf_ijcnn_LiangC08} for training and testing.
\end{itemize}

\textbf{SPML methods}:
We apply three methods that are based on SPML. Compared to SVM-based methods, these methods explicitly model the network structure information. Therefore, the  feature used here is simply the node attributes and links become linear constraints. Given an edge, its distance is just the Mahalanobis distance defined by learned metrics. The following methods are included:
\begin{itemize}
\item \textbf{ST-SPML}: This is the single-task SPML~\cite{ShawHuangJebara2011}. A metric is learned for each network independently. It models network structure but not task correlations.
\item \textbf{U-SPML}: ``U" means data pooling. Training examples from all tasks are pooled together and the learning procedure is simply ST-SPML. This is a naive way of sharing knowledge between tasks, but it does not respect the differences between and distinctiveness of tasks. Thus we expect inferior results, particularly for less related tasks.
\item \textbf{MT-SPML}: This is our method. By comparing it to other methods, we can demonstrate the fact that MT-SPML not only models the structure of all networks nicely, but also exploits  relatedness while respecting the distinctiveness of tasks.
\end{itemize}
\begin{table}
\begin{center}
\begin{tabular}{|c|c|c|c|}
\hline
Areas & \# of nodes & \# of edges & \# of features\\
\hline
Search Engine & 269 & 332 & 6695\\
\hline
Graph Theory & 223 & 917 & 6695\\
\hline
Philosophy & 303 & 921 & 6695\\
\hline
\end{tabular}
\caption {Statistics of Wikipedia article data}
\end{center}
\end{table}

Finally, we also compare to the direct use of the original feature vector, i.e., using Euclidean distance between feature vectors as the distance. While we are aware of the existence of other link prediction methods, such as Adamic-Adar~\cite{Liben_Nowell_2003}. As we have already mentioned, Adamic-Adar~\cite{Liben_Nowell_2003} only predicts potential links for nodes that are already present in the network and thus rely heavily on a snapshot of dense network structure. Thus, it is not suitable for our experiments, where no initial links for the test node are provided. Moreover, CNLP~\cite{icde_QiAH13} targets at improving link prediction on one specific network by using other networks. Thus, its goal is fundamentally different from ours. Our methods encode observed structures in the learned metric and can be used for both unobserved test nodes and sparse graphs.

 All the results are reported in Fig.1. The first thing we see is that  SVM-based  methods perform the worst when there are fewer training examples while the SPML family achieves good results in all settings, due to its ability to model structure information. We also find that among the SPML methods, MT-SPML consistently outperforms the others, which implies that MT-SPML is better at exploiting task correlations. Interestingly, we notice that the least amount of improvement from MT-SPML is found for philosophy articles. This observation seems to be aligned with the intuition that search engine-related  and graph theory-related papers probably have more in common with each other than with philosophy papers. 

We also show the convergence behavior of MT-SPML by plotting the value of $|S_{q+}|$,  number of violated constraints among those randomly sampled ones, for every iteration for each task. The fewer the number of violated constraints, the better the new metric respects the network structure. In experiments we set $B$, the time of random sampling, to be $10$. In order to make a clearer demonstration, in Fig.2 we set $B$ to be $100$. As shown by Fig.2, the numbers of violated constraints of all tasks drop quickly within the first $1000$ iterations and stabilizes after $4000$ iterations.
%
%\begin{figure*}
%\begin{minipage}{10cm}
%\hspace{-26mm}\epsfig{file=Fig1/overall.eps, width=4.5in}\vspace{3mm}
%\end{minipage}\\
%\hspace{5mm}
%\begin{minipage}{15cm}
%\hspace{-20mm}\epsfig{file=Fig1/searchengine.eps, width=2in}
%\hspace{-8mm}\epsfig{file=Fig1/graphtheory.eps, width=2in}
%\epsfig{file=Fig1/philosophy.eps, width=2in}
%\end{minipage}
%\caption{Link prediction performance on Wikipedia article data. Smaller figures in the lower are AUC numbers with training sets of various sizes for each area. The bigger figure on the top is the average AUC performances over all three areas. }
%\end{figure*}

\begin{figure*}
\centering
\hspace{-20mm}\begin{minipage}{10cm}
\epsfig{file=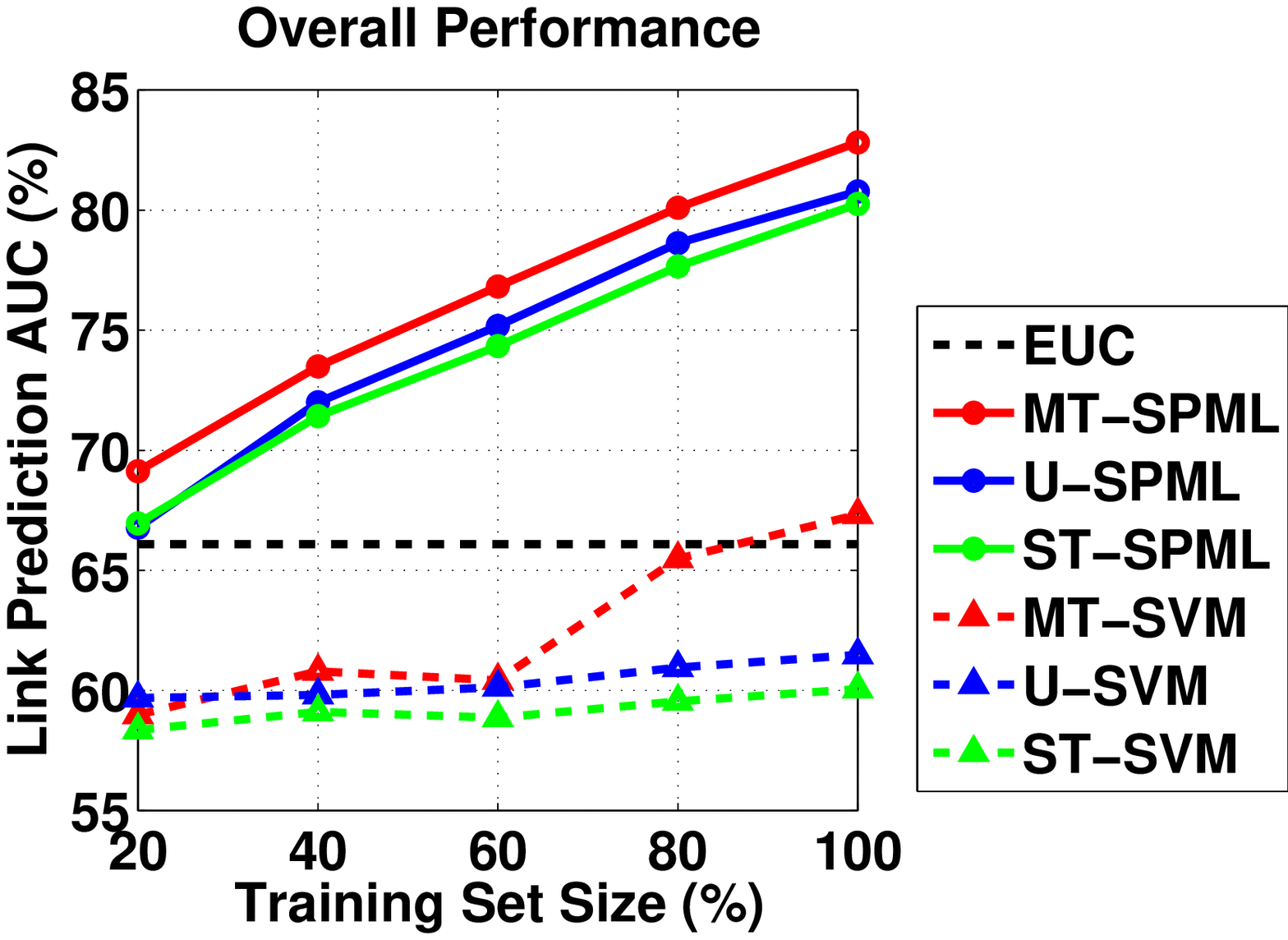, width=4.5in}\vspace{3mm}
\end{minipage}\\
\hspace{-5mm}\begin{minipage}{18cm}
\hspace{-3mm}\epsfig{file=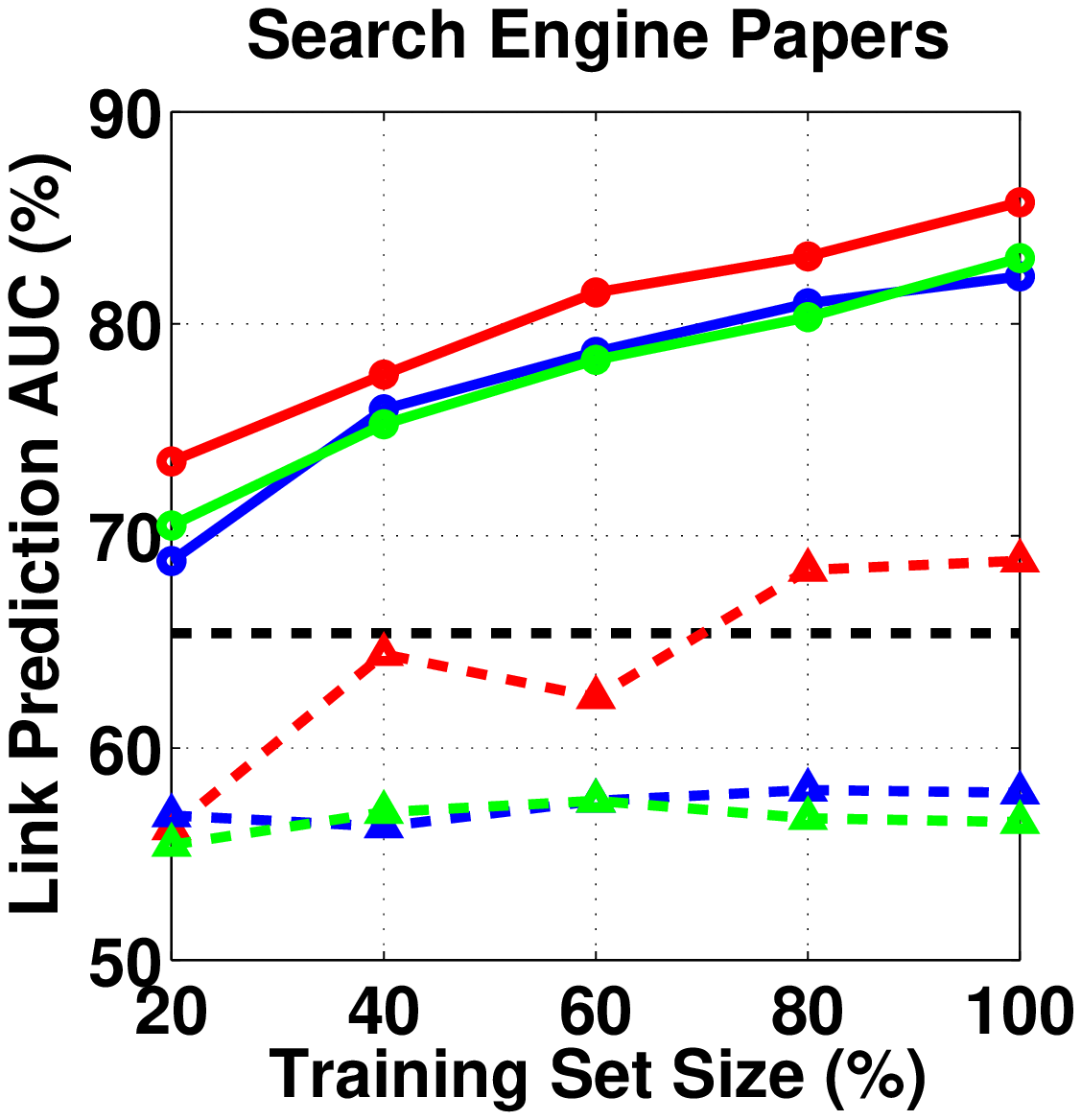, width=2.7in}
\hspace{-8mm}\epsfig{file=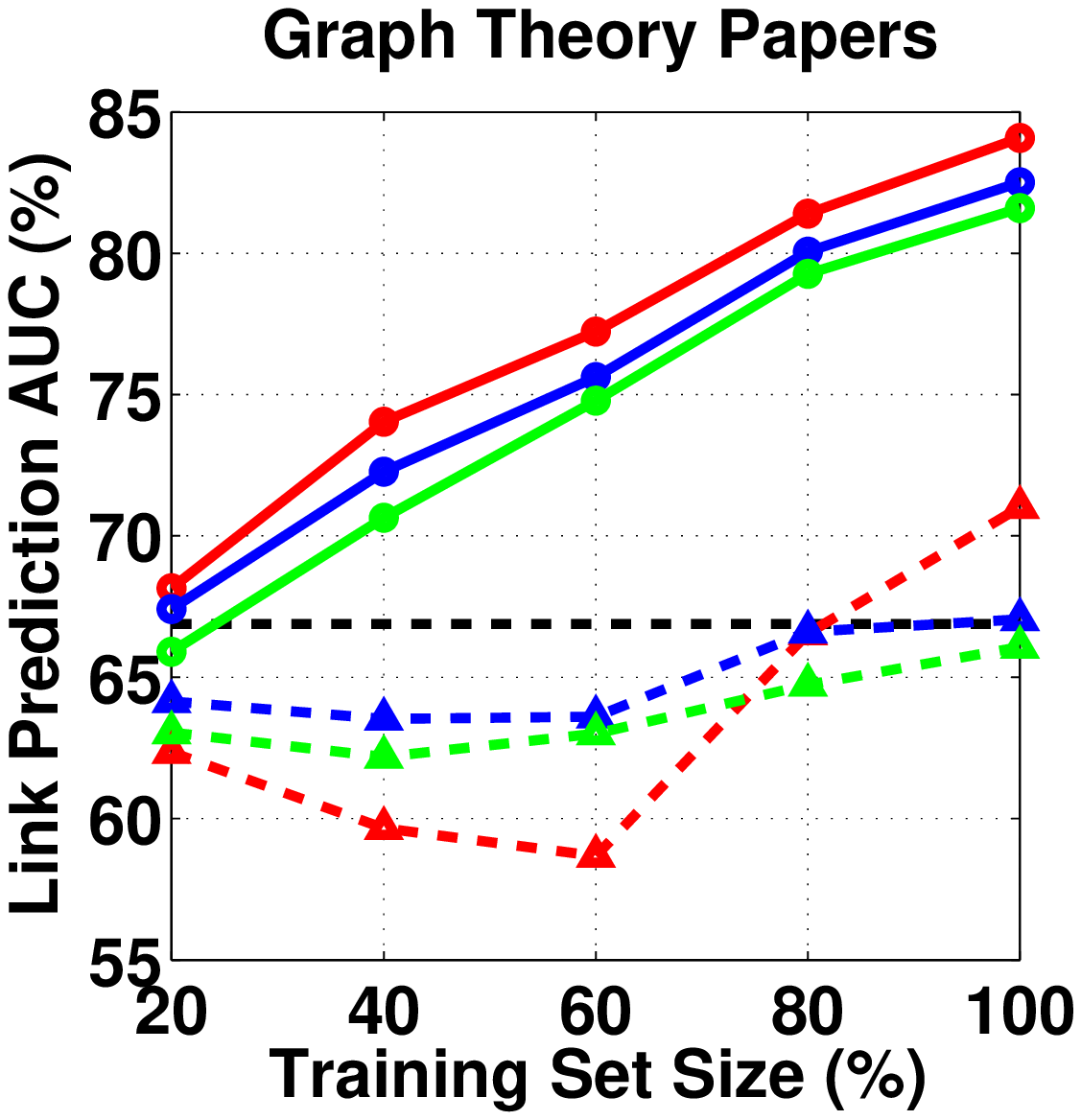, width=2.7in}
\hspace{-8mm}\epsfig{file=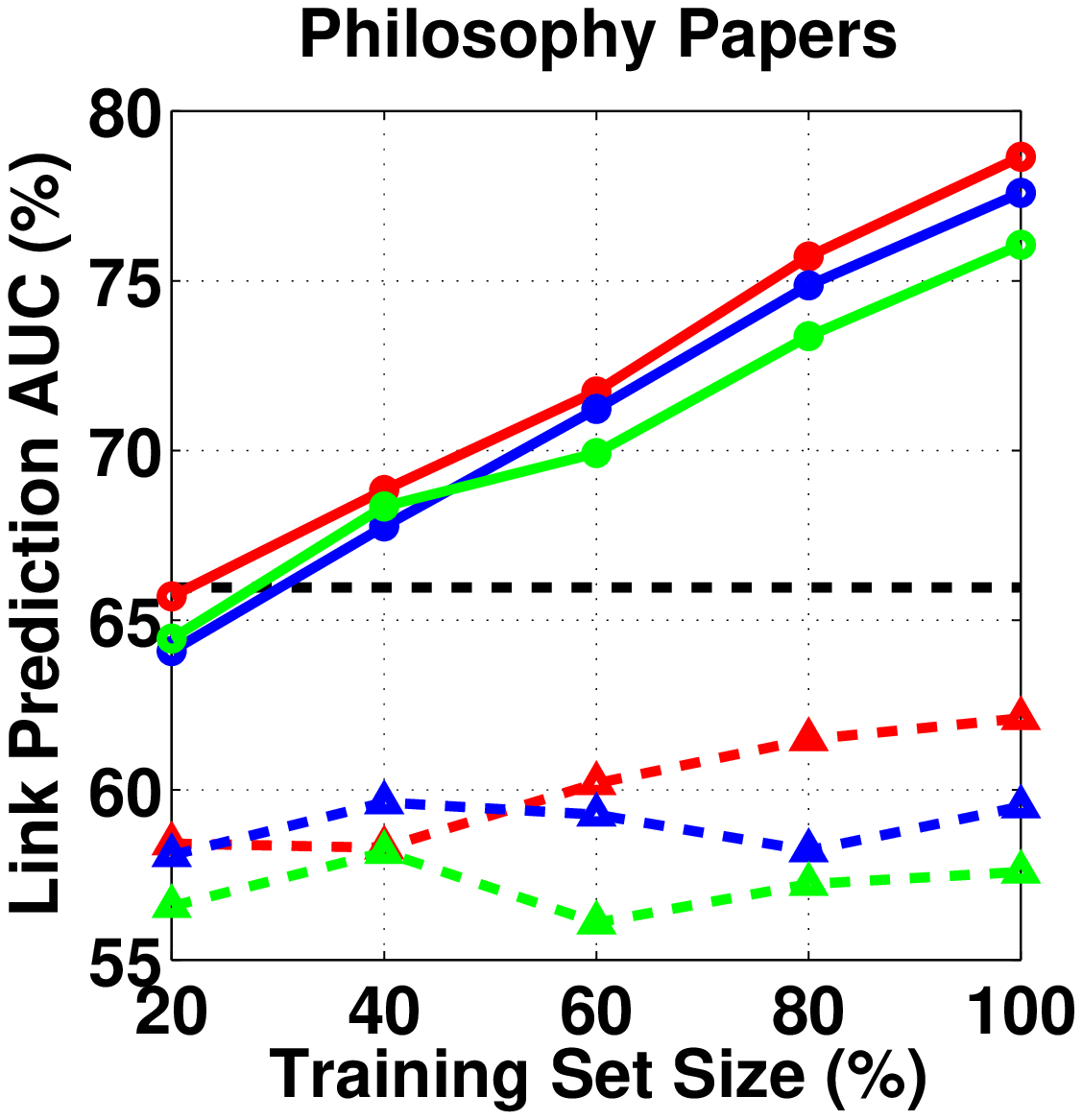, width=2.7in}
\end{minipage}
\caption{Link prediction performance on Wikipedia article data. Training set size is varied. Smaller figures in the lower half separate out the individual performance for each area. The bigger figure on the top is the average AUC performances over all three areas. }
\end{figure*}

\begin{figure*}
\centering
\epsfig{file=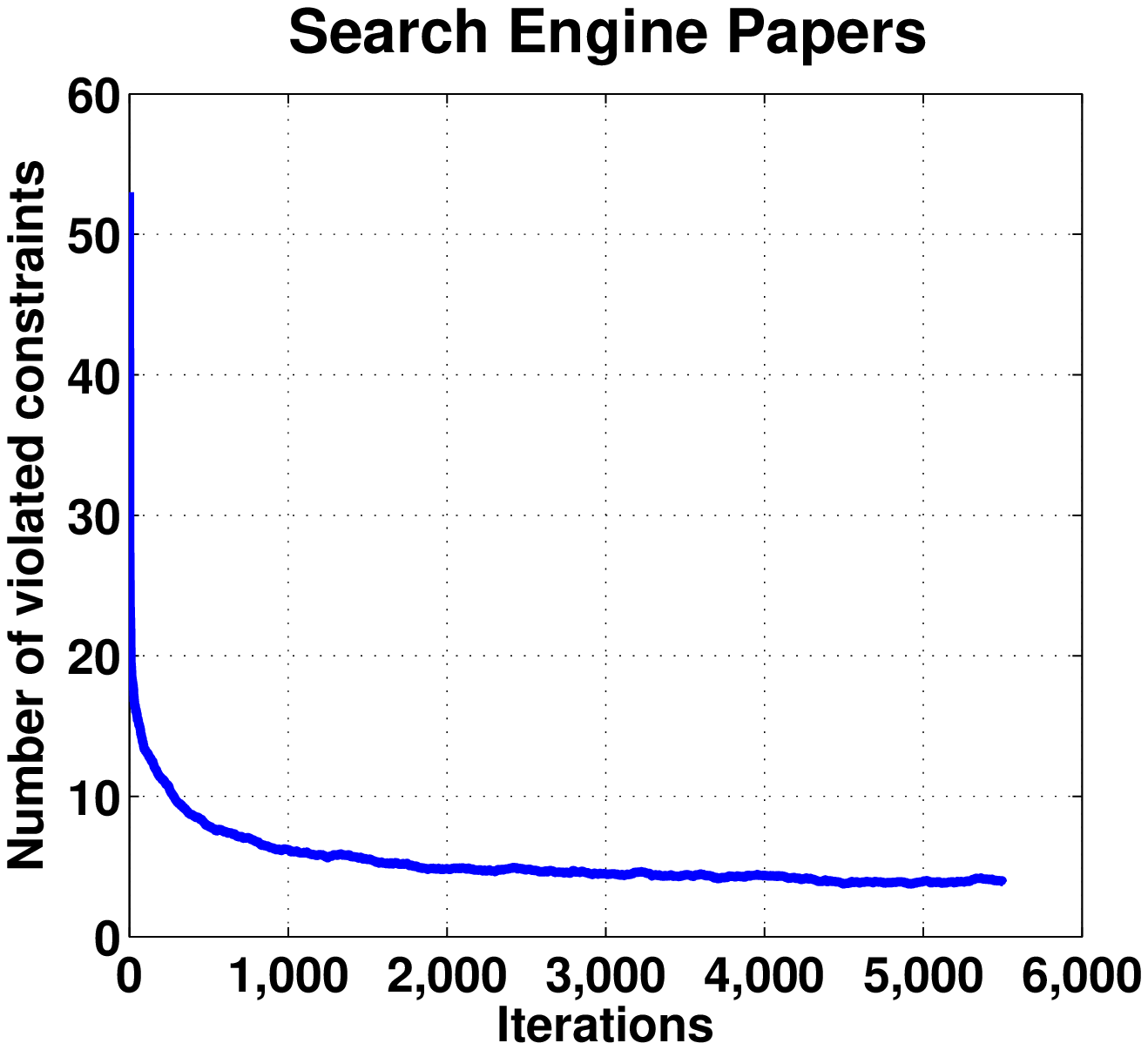, width=2.25in}
\epsfig{file=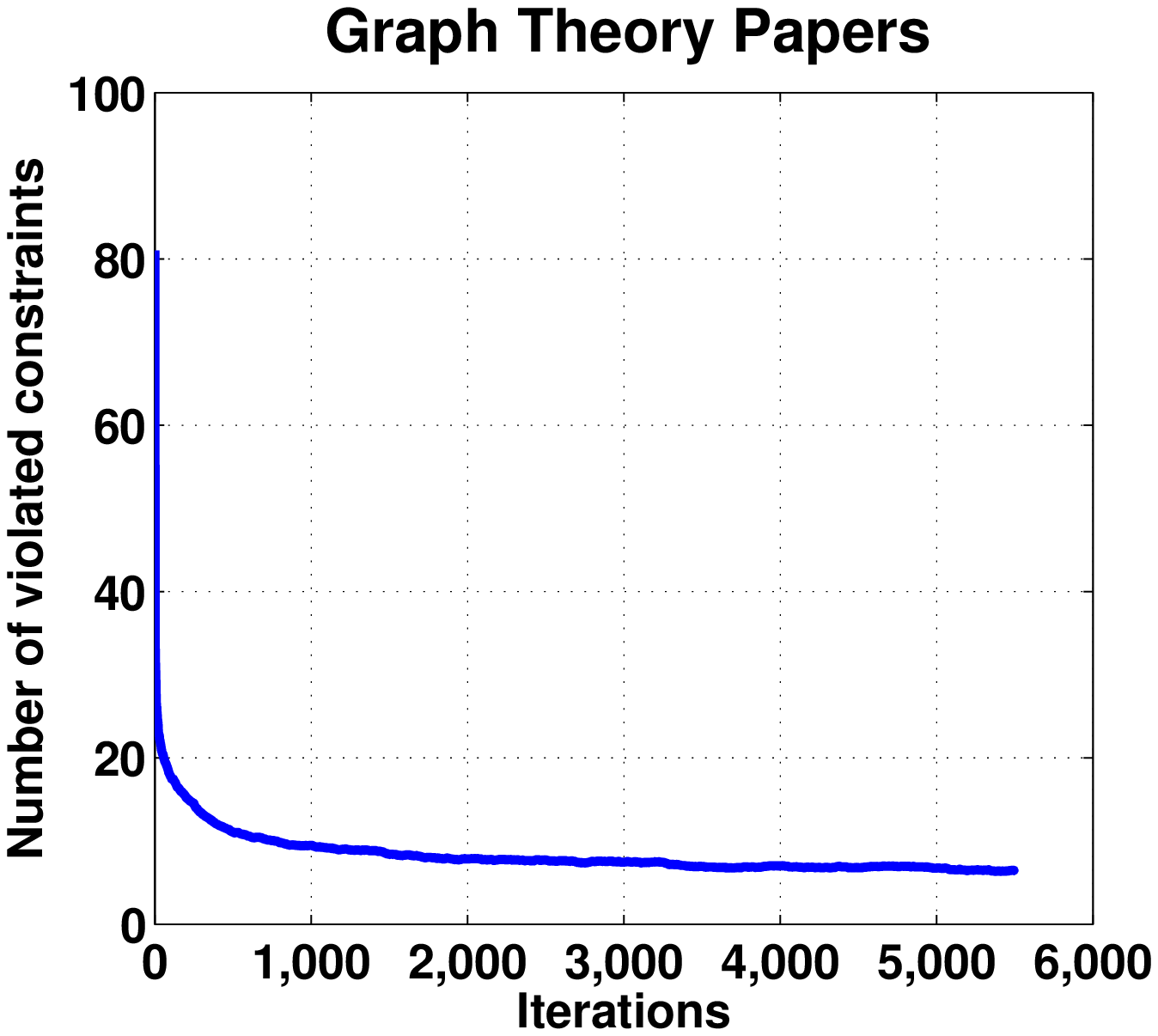, width=2.25in}
\epsfig{file=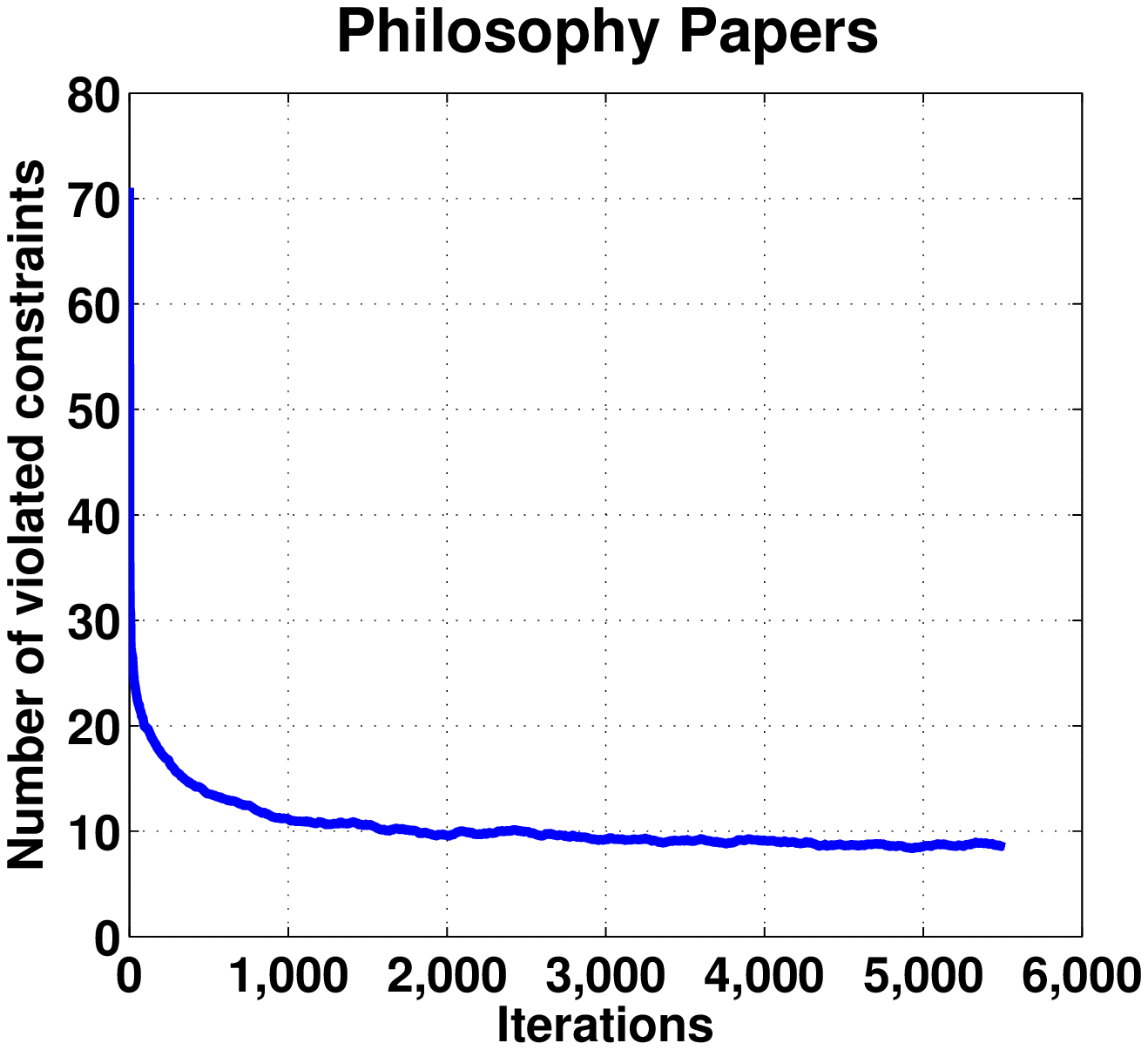, width=2.25in}
\caption{Number of violated constraints within first $5500$ iterations. $100$ constraints are sampled at every iteration.}
\end{figure*}

\subsection{Social circle prediction on Google+}
Every member of an online social network (e.g., Google+) is the ego of his/her (sub-)network and tends -- or may be forced -- to categorize his/her relationships (e.g. family members, college friends or childhood friends). For each class of relationships, there is a sub-network associate with it, the {\em social circle}, which is directly formalized in the online structures of  Google+ (see ~\cite{nips_McAuleyL12}). In this section, given a social network user (the ego) and his/her friends, we want to predict his/her social circles, namely the type of relationships between ego and ego's friends  based on profile information. We are only interested in the ego network, meaning that we do not predict the links between friends. A similar topic is studied by McAuley et al~\cite{nips_McAuleyL12}, where the setup is very different from ours. They assume the observation of an entire ego network, including attributes and structure, but not any social circle labels, and the goal is to assign social circle labels to links in an unsupervised manner. Our problem uses a supervised learning setting, where we observe only parts of the network and the corresponding social circle labels. For the prediction of each social circle, we treat it as link prediction. However, as mentioned in our introduction (Section 1), the correlation between social circles should be exploited. Thus, we treat the prediction of each social circle as a task, and MT-SPML is applied to learn metrics jointly over the underlying ego networks of all social circles. Note that, as reported in~\cite{nips_McAuleyL12}, social circles largely overlap with each other, which implies strong correlations and MTL is thus likely to achieve a more significant performance gain. We obtain data from~\cite{nips_McAuleyL12}, which was from Google+ users and information is anonymous. We randomly pick one user and his/her social circles for our experiment. The entire ego network has $4402$ nodes and $5$ social circles. The profile of all nodes is also preserved. There are $6$ classes of feature types, including gender, institution, job title, last name, place, and university. We build a bag-of-words feature for all feature types and concatenate them all, resulting in a feature vector of $2969$ dimensions.

In this experiment, we adopt a different procedure. We start with using ST-SPML to learn a metric for each social circle independently. Then, to show the advantage of doing MTL jointly over multiple tasks, we run MT-SPML on various numbers of social circles.  To avoid the exponential number of social circle combinations, we index them from 1 to 5. We begin by running on \textbf{ \{1,2\}} and add one more social circle at a time in order, resulting in the following four combinations: \textbf{ \{1,2\}, \{1,2,3\}, \{1,2,3,4\}, \{1,2,3,4,5\} } which we will continue to use in the following experiments. In this way, we can see the behavior of the algorithms as more relevant tasks joining. In Fig. 3, we compare ST-SPML to MT-SPML on the four combinations of social circles. Note that, because of the inferior performance of SVM based methods on Wikipedia article data, we entirely omit them in this experiment. Clearly, as shown in Fig. 3, all social circles benefit from MTL and the improvement is significant, except for Social Circle 2, whose performance gain is slight. We speculate that Social Circle 2 is not closely related to other circles (e.g., in terms of the number of overlapping nodes). We will discuss the case of Social Circle 2 later.

Now we compare MT-SPML to U-SPML, which simply pools all data together and estimates a model for all tasks. Both MT-SPML and U-SPML are applied to the four combination settings of different social circles. As shown by Fig. 4, MT-SPML consistently and significantly outperforms U-SPML at all locations. 

Now we would like to further investigate  Social Circle 2. We first show some statistics in Table 2, where we show the percentage of node overlapping between each pair of social circles. The overlap is defined as the intersection of nodes over the union. As we can see, some circles are largely overlapped (e.g., \textbf{\{1,3\}} have 81.9\% nodes in common), while Social Circle 2 barely overlaps with the others. Although overlapping is not the only quantitative measurement of correlations between social circles,  a substantial set of common nodes suggests that there are some shared semantics between two relationships. The statistics of Table 2 supports our earlier speculation as to why Social Circle 2 does not benefit from joint learning as much as the others. 

Furthermore, we would like to again show the advantage of MT-SPML by showing the results of a pair of tasks that are less correlated to each other. We choose Social Circles \textbf{\{1,2\}}, since they have only 1.1\% nodes in common. In Fig. 5, MT-SPML is jointly learned on \textbf{\{1,2\}}, U-SPML is learned via data pooling, and ST-SPML is trained on 1 and 2 independently. The prediction performances of two tasks are reported in the two groups of bars respectively. As shown in Fig. 5, MT-SPML still gets 2\%-5\% performance improvement over ST-SPML (bars with circles on top). However, the strategy of simple data pooling used by U-SPML (bars with down pointing triangle) reduces the performance (produces results worse than ST-SPML). This observation suggests that on difficult cases where tasks are less relevant, MTL is still able to utilize useful correlations, while respecting the boundaries between tasks.

\begin{figure*}
\centering
\hspace{-8mm}\epsfig{file=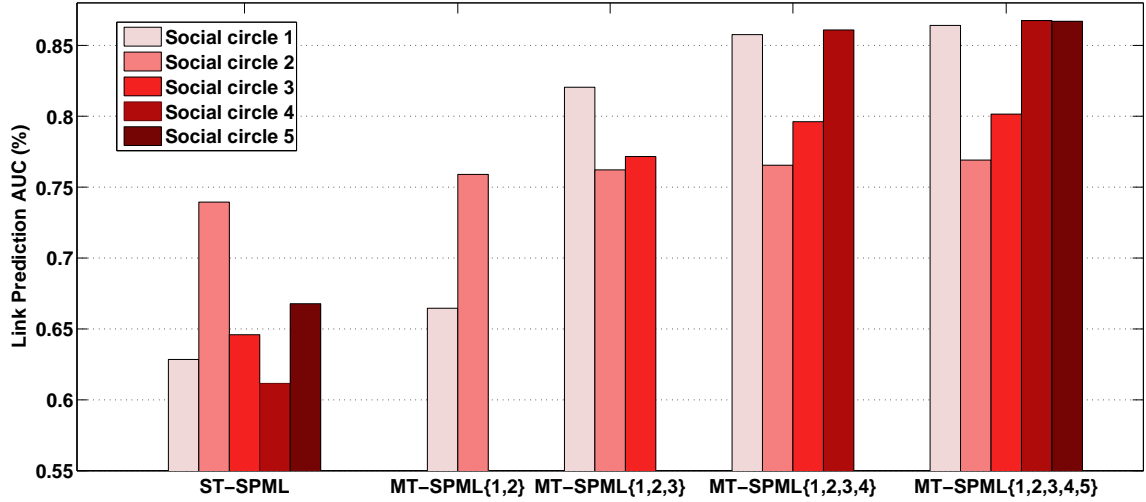, width=7.2in}
\caption{Link prediction performance on Google+ data. Social circles are color coded. The comparison is between ST-SPML and MT-SPML. The first group contains the prediction performance of ST-SPML on all social circles, while the other groups show the performance of MT-SPML that learned and tested on multiple combinations of social circles, for example, \textbf{MT-SPML\{1,2,3\}} means learning and testing on Social Circle 1, 2 and 3.}
\end{figure*}

\begin{figure*}
\centering
\hspace{-10mm}\epsfig{file=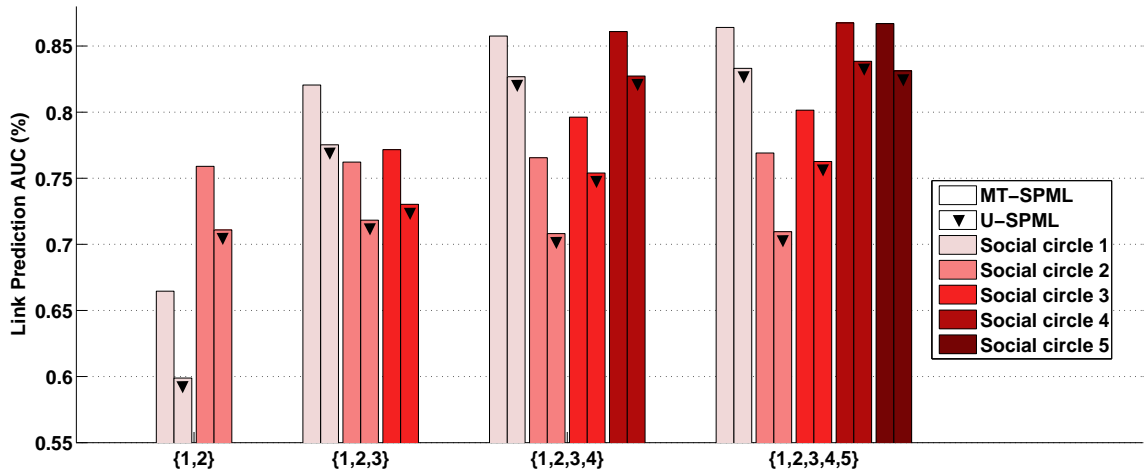, width=7.2in}
\caption{Link prediction performance on Google+ data. The comparison is between MT-SPML and U-SPML. Social circles are color coded. Different methods for the same task are compared side by side. U-SMPL is indicated by a downward pointing triangle. Each group is trained and tested on a set of social circles. For example, \textbf{\{1,2,3\}} means learning and testing on Social Circle 1, 2 and 3.}
\end{figure*}

\begin{table}
\begin{center}
\begin{tabular}{|c|c|c|c|c|c|}
\hline
Social Circles&1&2&3&4&5\\
\hline
1&0&1.1\%&81.9\%&89.6\%&84.1\%\\
\hline
2&1.1\%&0&0.9\%&1.1\%&1.1\%\\
\hline
3&81.9\%&0.9\%&0&73.5\%&68.9\%\\
\hline
4&89.6\%&1.1\%&73.5\%&0&93.7\%\\
\hline
5&84.1\%&1.1\%&68.9\%&93.7\%&0\\
\hline
\end{tabular}
\caption {Statistics of overlapping nodes between social circles. Overlapping ratios are presented.}
\end{center}
\end{table}

\begin{figure}
\centering
\epsfig{file=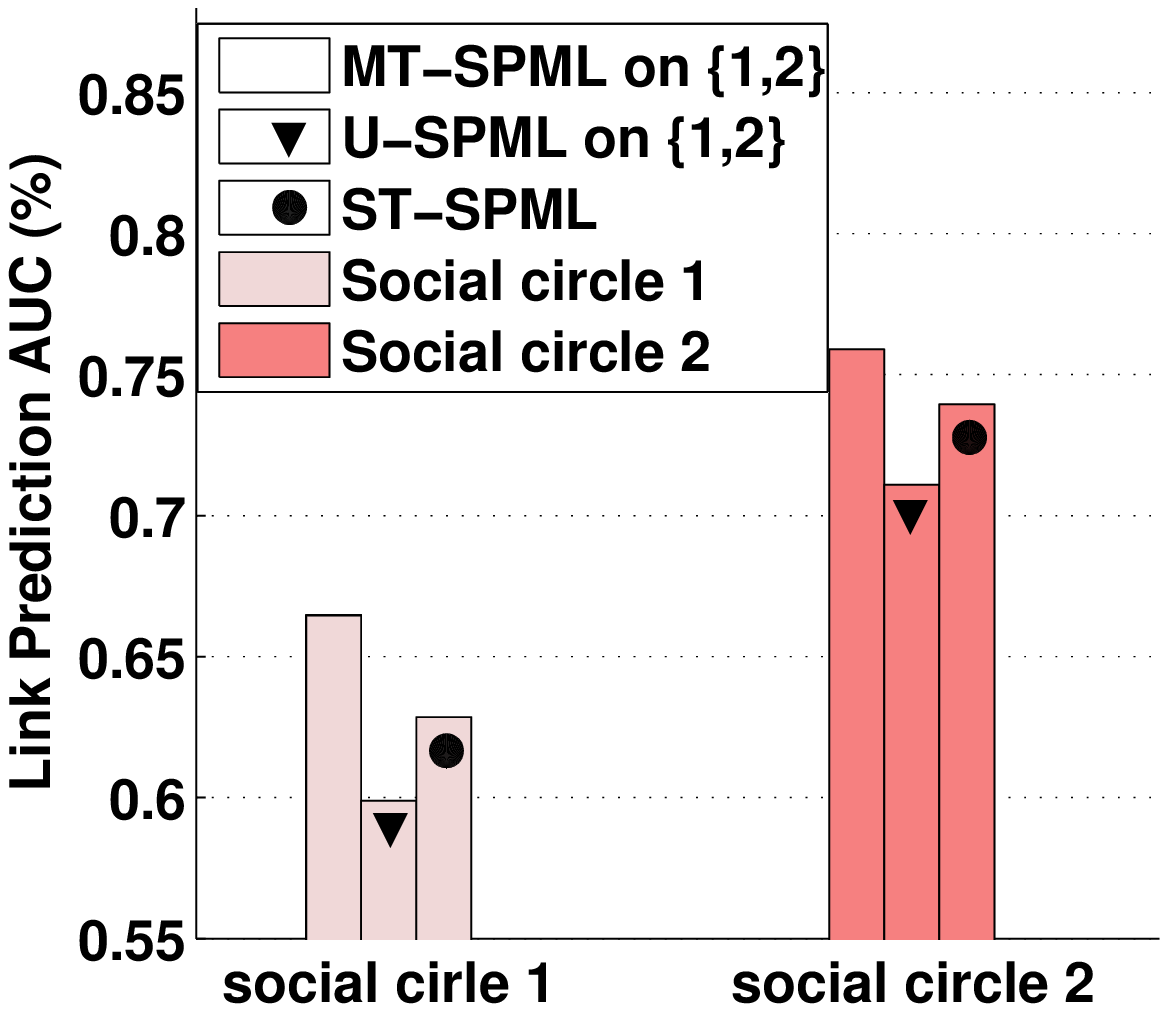, width=3.2in}
\caption{U-SPML negatively impacts performance when training on \{1,2\}, two less relevant tasks. MT-SPML is able to improve performance compared to ST-SPML by exploiting useful correlations, while U-SPML gets inferior results.}
\end{figure}
\section{Conclusions}
In this paper, we proposed MT-SPML, a large margin-based multi-task learning method for network data. It operates on networks with node attributes. It learns a task specific distance metric for every task and a common distance metric for all. By combining a task specific metric with the common distance, the final metric preserves the structure of the corresponding network, thus it can be used to predict link patterns on sparse nascent networks or for incoming nodes at test time. We applied MT-SPML to two common real-world problems, article citation prediction and social circle prediction. Better results were achieved (as compared to reasonable baselines) and detailed analysis was provided. The importance of our work lies in the fact that network data has large variation and diversity, thus many related tasks can be performed, and we are able to better exploit the useful correlations. Moreover, since MT-SPML is a general method and can be optimized via stochastic gradient descent with good convergence behavior, it is suitable for general (and large) network data and can be widely applied to real-world problems. All the code and data used in experiments are available for download at the link given in the paper.\\
%\end{document}  % This is where a 'short' article might terminate

%ACKNOWLEDGMENTS are optional
\section{Acknowledgments}
We are grateful to our funding sources. The authors were supported by AFOSR Award FA9550-11-1-0166 and the Neukom Institute for Computational Science.

%
% The following two commands are all you need in the
% initial runs of your .tex file to
% produce the bibliography for the citations in your paper.
\bibliographystyle{abbrv}
\bibliography{sigproc}  % sigproc.bib is the name of the Bibliography in this case

\begin{thebibliography}{10}

\bibitem{Adafre:2005:DML:1134271.1134284}
S.~F. Adafre and M.~de~Rijke.
\newblock Discovering missing links in wikipedia.
\newblock In {\em Proceedings of the 3rd International Workshop on Link
  Discovery}, LinkKDD '05, 2005.

\bibitem{Agarwal_2010}
A.~Agarwal, H.~Daume~III, and S.~Gerber.
\newblock Learning multiple tasks using manifold regularization.
\newblock In {\em NIPS}. 2010.

\bibitem{ShawHuangJebara2011}
B.~H. Blake~Shaw and T.~Jebara.
\newblock Learning a distance metric from a network.
\newblock In {\em NIPS}, 2011.

\bibitem{Caruana97multitasklearning}
R.~Caruana.
\newblock Multitask learning.
\newblock In {\em Machine Learning}, pages 41--75, 1997.

\bibitem{ChechikSSB10}
G.~Chechik, V.~Sharma, U.~Shalit, and S.~Bengio.
\newblock Large scale online learning of image similarity through ranking.
\newblock {\em Journal of Machine Learning Research}, 11:1109--1135, 2010.

\bibitem{cikm_ComarTJ10}
P.~M. Comar, P.-N. Tan, and A.~K. Jain.
\newblock Multi task learning on multiple related networks.
\newblock In {\em CIKM}, 2010.

\bibitem{Daume2009BML}
H.~Daum{\'e}, III.
\newblock Bayesian multitask learning with latent hierarchies.
\newblock In {\em UAI}, 2009.

\bibitem{Evgeniou_2004}
T.~Evgeniou and M.~Pontil.
\newblock Regularized multi--task learning.
\newblock In {\em KDD}, 2004.

\bibitem{journals_jmlr_FanCHWL08}
R.-E. Fan, K.-W. Chang, C.-J. Hsieh, X.-R. Wang, and C.-J. Lin.
\newblock Liblinear: A library for large linear classification.
\newblock {\em Journal of Machine Learning Research}, 9:1871--1874, 2008.

\bibitem{Hasan06linkprediction}
M.~A. Hasan, V.~Chaoji, S.~Salem, and M.~Zaki.
\newblock Link prediction using supervised learning.
\newblock In {\em In Proc. of SDM 06 workshop on Link Analysis,
  Counterterrorism and Security}, 2006.

\bibitem{4053061}
H.~Kashima and N.~Abe.
\newblock A parameterized probabilistic model of network evolution for
  supervised link prediction.
\newblock In {\em ICDM}, 2006.

\bibitem{conf_ijcnn_LiangC08}
L.~Liang and V.~Cherkassky.
\newblock Connection between svm+ and multi-task learning.
\newblock In {\em IJCNN}, 2008.

\bibitem{Liben_Nowell_2003}
D.~Liben-Nowell and J.~Kleinberg.
\newblock The link prediction problem for social networks.
\newblock In {\em CIKM}, 2003.

\bibitem{journals/corr/abs-1303-1733}
B.~London, T.~Rekatsinas, B.~Huang, and L.~Getoor.
\newblock Multi-relational learning using weighted tensor decomposition with
  modular loss.
\newblock {\em CoRR}, abs/1303.1733, 2013.

\bibitem{nips_McAuleyL12}
J.~J. McAuley and J.~Leskovec.
\newblock Learning to discover social circles in ego networks.
\newblock In {\em NIPS}, 2012.

\bibitem{parameswaran10large}
S.~Parameswaran and K.~Weinberger.
\newblock Large margin multi-task metric learning.
\newblock In J.~Lafferty, C.~K.~I. Williams, J.~Shawe-Taylor, R.~Zemel, and
  A.~Culotta, editors, {\em NIPS}. 2010.

\bibitem{icde_QiAH13}
G.-J. Qi, C.~C. Aggarwal, and T.~S. Huang.
\newblock Link prediction across networks by biased cross-network sampling.
\newblock In {\em ICDE}, 2013.

\bibitem{Seltzer_deep2013}
M.~Seltzer and J.~Droppo.
\newblock Multi-task learning in deep neural networks for improved phoneme
  recognition.
\newblock In {\em Acoustics, Speech and Signal Processing (ICASSP), 2013 IEEE
  International Conference on}, 2013.

\bibitem{Shalev_ShwartzSSC11}
S.~Shalev-Shwartz, Y.~Singer, N.~Srebro, and A.~Cotter.
\newblock Pegasos: primal estimated sub-gradient solver for svm.
\newblock {\em Math. Program.}, 127(1):3--30, 2011.

\bibitem{icdm_TangLD09}
W.~Tang, Z.~Lu, and I.~S. Dhillon.
\newblock Clustering with multiple graphs.
\newblock In {\em ICDM}, 2009.

\bibitem{Taskar03linkprediction}
B.~Taskar, M.~fai Wong, P.~Abbeel, and D.~Koller.
\newblock Link prediction in relational data.
\newblock In {\em NIPS}, 2003.

\bibitem{Wang_cvpr09}
X.~Wang, C.~Zhang, and Z.~Zhang.
\newblock Boosted multi-task learning for face verification with applications
  to web image and video search.
\newblock In {\em CVPR}, 2009.

\bibitem{WeinbergerS09}
K.~Q. Weinberger and L.~K. Saul.
\newblock Distance metric learning for large margin nearest neighbor
  classification.
\newblock {\em Journal of Machine Learning Research}, 10, 2009.

\bibitem{Yu_2005}
K.~Yu, V.~Tresp, and A.~Schwaighofer.
\newblock Learning gaussian processes from multiple tasks.
\newblock In {\em ICML}, 2005.

\bibitem{ZhouZYSTZG08}
D.~Zhou, S.~Zhu, K.~Yu, X.~Song, B.~L. Tseng, H.~Zha, and C.~L. Giles.
\newblock Learning multiple graphs for document recommendations.
\newblock In {\em WWW}, 2008.

\end{thebibliography}
\end{document}